\documentclass[onecollumn,11pt]{article}
\usepackage[pdftex]{color,graphicx}
\usepackage{subfigure}
\usepackage{url}
\begin{document}
\DeclareGraphicsExtensions{.jpg,.pdf,.mps,.png}

\title{DimReduction - Interactive Graphic Environment for Dimensionality Reduction}

\date{}

\maketitle

\begin{center}

{\large
Fabr\'{i}cio Martins Lopes$^{(1,2)}$, David Correa Martins-Jr$^{(1)}$
and Roberto M. Cesar-Jr$^{(1)}$ \\
}

$^{(1)}$Instituto de Matem\'{a}tica e Estat\'{i}stica da Universidade de S\~{a}o Paulo, Brazil. \\
$^{(2)}$Universidade Tecnol\'{o}gica Federal do Paran\'{a}, Brazil.

\end{center}


\begin{abstract}
Feature selection is a pattern recognition approach to choose
important variables according to some criteria to
distinguish or explain certain phenomena. There are many genomic and proteomic
applications which rely on feature selection to answer questions
such as: selecting signature genes which are informative about some biological state, e.g.
normal tissues and several types of cancer; or defining a
network of prediction or inference among elements such as genes, proteins,
external stimuli and other elements of interest. In these
applications, a recurrent problem is the lack of samples to perform an
adequate estimate of the joint probabilities between element
states. A myriad of feature selection algorithms and
criterion functions are proposed, although it is difficult to point
the best solution in general. The intent of this work is to provide
an open-source multiplataform graphical environment to apply, test and compare
many feature selection approaches suitable to be used in bioinformatics problems.
\end{abstract}
\section{Introduction}
The pattern recognition methods allow the classification
of objects or patterns in a number of classes
\cite{theodoridis99}. Specifically in statistical pattern
recognition, given a set $Y = \{y_1, ..., y_c\}$ of
classes and an unknown pattern $\mathbf{X} = \{X_1, X_2, ...,X_n\}$, a
pattern recognition system associates $\mathbf{x}$ to a
class $y_i$ based on defined measures in a feature space. In many
applications, especially in bioinformatics, the feature space
dimension tends to be very large, making difficult the
classification task. In order to overcome this inconvenient situation,
the study of dimensionality reduction problem in pattern recognition
becomes imperative.

The so called ``curse of dimensionality'' \cite{jaduma:2000} is a
phenomenon in which the number of training samples
required to a satisfactory classifier performance is given by an
exponential function of the feature space. This is the main motivation
by which performing of dimensionality reduction is important in
problems with large number of features and small number of training
samples. Many bioinformatics applications are perfectly inserted in
this context. Data sets containing mRNA transcription expressions from
microarray or SAGE, for example, possess thousands of genes
(features) and only some dozens of samples that may be cell states or types
of tissues. If time is a factor involved, the samples are called
dynamical states, otherwise they are called steady states.

There are basically two dimensionality reduction approaches: feature extraction
and feature selection \cite{theodoridis99, duda00, camposdiss01}. The feature extraction methods create new
features from transformations or combinations of the original feature
set. On the other hand, feature selection algorithms just search for the
optimal feature subset according to some criterion function. The
software proposed in this paper is initially focused on feature
selection methods.

A feature selection method is composed by two main parts: a search
algorithm and a criterion function. As far as the search
algorithms, there are two main categories: the optimal and sub-optimal
algorithms. The optimal algorithms (including exhaustive and
branch-and-bound searches) return the best feature subspace,
but their computational costs are very high to be applied in
general. The sub-optimal algorithms do not guarantee that the solution
is optimal, but some of them present a reasonable cost-benefit
between computational cost and quality of the solution. Up to now, we
have implemented in the software the exhaustive search (optimal), the
Sequential Forward Selection (SFS - sub-optimal) and the Sequential
Forward Floating Selection (SFFS - sub-optimal with excellent cost-benefit)
~\cite{pudil94floating}.

There is a large number of criterion functions proposed in the literature.
The most common functions are based
on the classifier error and distances between patterns. There are also
criterion functions based on information theory. They are closely related
to the classifier error, but instead of using the error, it is based
on the conditional entropy of the class probabilities distributions
given the observed pattern.

Due to the curse of dimensionality phenomenon, error estimation is a
crucial issue. We have developed some ways to embed error estimation
in the criterion functions based on classifier error or conditional
entropy. The main idea is based on penalization of non-oberved or rarely
observed instances. A good advantage in doing this is that the right
dimension of the feature subset solution is also estimated (the
dimension parameter is not required). After the feature selection, it
is possible to apply classical error estimation techniques like
resubstitution, leave-one-out, cross validation or bootstrap.

The software is implemented in Java, so it can be executed in many
operational systems. It is open source and intended to be continuously
developed in a world-wide collaboration. The software is available at
\url{http://code.google.com/p/dimreduction/}.

Following this introduction, Section~\ref{sec:feature_selection} and
\ref{sec:criterion_functions} will describe the feature selection
algorithms and criterion functions implemented so far.
Section~\ref{sec:software} discusses the implemented software.
Section~\ref{sec:results} will shows some preliminary results obtained on gene
regulation networks and classification of breast cancer cells. This paper is finalized with some conclusions in
Section~\ref{sec:conclusion}.
\section{Implemented feature selection algorithms}
\label{sec:feature_selection}

The first and simpler feature selection algorithm implemented in this
work is the exhaustive search. This algorithm searches the whole search space,
and as a result, the selected features are optimal. However in bioinformatics
context, normally the computational cost makes this approach
inadequate. Then, it is clear the existence of a trade-off between
optimality and computational cost.

An alternative way is to adopt sub-optimal search
methods. In this work we have implemented two sub-optimal approaches with
unique solution, which are known as top down and bottom up.
In the first one, the selection subset starts empty and features are
inserted by optimizing a criterion function until a stop condition is satisfied,
which is often based on the subset size or a threshold. In the second algorithm,
the subset starts full and features are removed, trying to optimize the criterion
function until a stop condition is reached. Methods that implement these approaches
are known as SFS (Sequential Forward Search) and SBS (Sequential Backward Search),
respectively. Considering the context of this work, our choice was to implement the SFS approach.

However, these suboptimal search methods present an undesirable
drawback known as nesting effect. This effect happens because the
discarded features in the top-down approach are not inserted anymore,
or the inserted features in the bottom-up approach are never discarded.

In order to circumvent this problem, the Sequential Forward Floating
Selection (SFFS)~\cite{pudil94floating} was also implemented. The SFFS
algorithm tries to avoid the nesting effect allowing to insert and exclude
features on subset in a floating way, i.e. without defining the number of
insertions or exclusions.

The SFFS may be formalized as in~\cite{pudil94floating}.
Let $\mathbf{X_k} = \{x_i: 1 \leq i \leq k, x_i \in \mathbf{X}\}$ be
the subset with $k$ features of the complete set $\mathbf{X} =
\{x_i: 1 \leq i \leq n\}$ with $n$ features
available. Let $E(\mathbf{X_k})$ the criterion function value for the subset
$\mathbf{X_k}$. The algorithm initializes with $k=0$, therefore the
subset $\mathbf{X_k}$ is empty.

First Step (insert): using the SFS method, select the feature $x_{k+1}$ of
the set $\mathbf{X} - \mathbf{X_k}$ to form the set
$\mathbf{X_{k+1}}$, such that
$x_{k+1}$ be the most relevant feature of the subset $\mathbf{X_k}$. The
new subset is $\mathbf{X_{k+1}} = \mathbf{X_k} \cup x_{k+1}$.

Second Step (conditional exclusion): Find the least relevant feature
in the set $\mathbf{X_{k+1}}$. If $x_{k+1}$ is the least relevant
feature in the subset $\mathbf{X_{k+1}}$, then $k \leftarrow k+1$,
$\mathbf{X_k} \leftarrow \mathbf{X_{k+1}}$ and back to the first
step. If $x_r, 1 \leq r \leq k$ is the least relevant feature in the
subset $\mathbf{X_{k+1}}$, then exclude $x_r$ from $\mathbf{X_{k+1}}$
to form a new subset $\mathbf{X'_k} = \mathbf{X_{k+1}} - x_r$ and
$k \leftarrow k-1$. If $k = 2$, then $\mathbf{X_k} =
\mathbf{X'_k}$, and return to the first step, else execute the third step.

Third Step (continuation of conditional exclusion): Find the least
relevant feature $x_s$ in the set $\mathbf{X'_k}$. If
$E(\mathbf{X'_k}- x_s) \leq E(\mathbf{X_{k-1}})$, then
$\mathbf{X_k} \rightarrow \mathbf{X'_k}$ and return to first step. If
$E(\mathbf{X'_k}-x_s) > E(\mathbf{X_{k-1}})$ then exclude $x_s$ from
$\mathbf{X'_k}$ to form a new reduced subset
$\mathbf{X'_{k-1}} = \mathbf{X'_k} - x_s$ and
$k \rightarrow k-1$. If $k = 2$, then $\mathbf{X_k} = \mathbf{X'_k}$
and return to first step, else repeat the third step.

The SFFS algorithm starts by setting $k=0$ e $\mathbf{X_k} = 0$, and the SFS
method is used until the subset size $k=2$.Then the SBS is performed
in order to exclude bad features. SFFS proceeds by alternating between SFS and
SBS until a stop criteria is reached. The best result set for each cardinality
is stored in a list. The best set among  them is selected as algorithm result,
and tie occurs, the set with lower cardinality is selected.
\section{Implemented criterion functions}
\label{sec:criterion_functions}

We implemented criterion functions based on classifier information
(mean conditional entropy) and
classifier error (Coefficient of Determination \cite{doughertykim}),
introducing some penalization on poorly or non-observed patterns.
\subsection{Mean conditional entropy}

The information theory was originated by Shannon
\cite{Shannon} and can be employed on feature selection problems
\cite{duda00}. The Shannon's entropy $H$ is a measure of randomness of
a variable $Y$ given by:

\begin{equation} \label{eq:h}
H(Y) = - \sum_{y \in Y}  P(y)  log  P(y) \textrm{,}
\end{equation}

\noindent where $P$ is the probability distribution function. By
convention $0 \cdot log 0 = 0$.

The conditional entropy is a fundamental concept
related to the mutual information. It is given by the following equation:

\begin{equation} \label{eq:ce}
H(Y|\mathbf{X} = \mathbf{x}) = - \sum_{y \in Y} P(y|\mathbf{X} = \mathbf{x})
log P(y|\mathbf{X} = \mathbf{x})
\end{equation}

\noindent where $\mathbf{X}$ is a feature vector and
$P(Y|\mathbf{X} = \mathbf{x})$ is the conditional
probability of $Y$ given the observation of an instance $\mathbf{x}
\in \mathbf{X}$. And finally, the mean conditional entropy of $Y$
given all the possible instances $\mathbf{x} \in \mathbf{X}$ is given by:

\begin{equation} \label{eq:mce}
H(Y|\mathbf{X}) = \sum_{\mathbf{x} \in \mathbf{X}} P(\mathbf{x})
H(Y|\mathbf{x})
\end{equation}

Lower values of $H$ yield better feature subspaces (the lower $H$,
the larger is the information gained about $Y$ by observing $\mathbf{X}$).
\subsection{Coefficient of Determination}

The Coefficient of Determinstion (CoD) \cite{doughertykim}, like
the conditional entropy, is a non-linear criterion useful for feature
selection problems \cite{Hsing}. It is given by:

\begin{equation} \label{eq:cod}
CoD_Y(\mathbf{X}) = \frac{1 - \max_{y \in Y}P(y) - (1 -
\sum_{\mathbf{x} \in \mathbf{X}} P(\mathbf{x})\max_{y \in
Y} P(y|\mathbf{x}))}{1 - \max_{y \in Y}P(y)}
\end{equation}

\noindent where $1 - \max_{y \in Y}P(y)$ is the error of predicting
$Y$ in the absence of other observations (let us denote it by
$\varepsilon_Y$) and $1 - \sum_{\mathbf{x} \in \mathbf{X}}
\max_{y \in Y}P(\mathbf{x},y)$ is the error of
predicting $Y$ based
on the observation of $\mathbf{X}$ (let us denote it by
$\varepsilon_Y(\mathbf{X})$). Larger values of $CoD$ yield
better feature subspaces ($CoD = 0$ means that the feature subspace
does not improve the priori error and $CoD = 1$ means that the error
was fully eliminated).
\subsection{Penalization of non-observed instances}

A way to embed the error estimation caused by using feature vectors
with large dimensions and insufficient number of samples is to
involve non-observed instances in the criterion value calculus
\cite{Martins_paa}. A
positive probability mass is attributed to the non-observed instances
and their contribution is the same as observing only the $Y$ values
with no other observations.

In the case of mean conditional entropy,
the non-observed instances get the entropy equal to $H(Y)$  and,
for the $CoD$, they get the prior error $\varepsilon_Y$ value. The
probability mass for the non-observed instances is parametrized by
$\alpha$. This parameter is added to the relative frequency (number of
occurrences) of all possible instances. So, the mean conditional
entropy with this type of penalization becomes:

\begin{equation} \label{eq:mce2}
H(Y|\mathbf{X}) = \frac{1}{\alpha M + s}\left[\alpha (M - N) H(Y) +
\sum_{i=1}^N (f_i + \alpha) H(Y|\mathbf{X} = \mathbf{x_i})\right]
\end{equation}

\noindent where $M$ is the number of possible instances of the feature
vector $\mathbf{X}$, $N$ is the number of observed instances (so, the
number of non-observed instances is given by $M - N$), $f_i$ is the
relative frequence (number of observations) of the instance
$\mathbf{x_i}$ and $s$ is the number of samples.

And $CoD$ becomes:

\begin{equation} \label{eq:cod2}
CoD_Y(\mathbf{X}) = \frac{\varepsilon_Y -
\left[\frac{\alpha (M - N)\varepsilon_Y}{\alpha M + s}
+ 1 - \sum_{i=1}^N\frac{(f_i + \alpha)}{\alpha M + s}
\max_{y \in Y}P(y|\mathbf{x_i})\right]}{\varepsilon_Y}
\end{equation}
\subsection{Penalization of rarely observed instances}

In this penalization, the non-observed instances are not taken into
account. This penalization consists in changing the conditional
probability distribution of the instances that have just a unique
observation \cite{barreranando}. It makes sense because if an instance
$\mathbf{x}$ has only 1 observation, the value of $Y$ is fully determined
($H(Y|\mathbf{X} = \mathbf{x}) = 0$ and $CoD_Y(\mathbf{X}) = 1$), but
the confidence
about the real distribution of $P(Y|\mathbf{X} = \mathbf{x})$ is very
low. A parameter $\beta$ gives a confidence value that $Y = y$.
The main idea is to distrubute $1 -\beta$ equally over all
$P(Y \neq y|\mathbf{X} = \mathbf{x})$ and to attribute $\beta$ to
$P(Y = y|\mathbf{X} = \mathbf{x})$. In Barrera \textit{et al}~\cite{barreranando},
the $\beta$ value is $\frac{1}{|Y|}$ where $|Y|$ is the number of
classes (cardinality of $Y$), becoming the uniform distribution (strongest penalization).

Adapting this penalization to the Equation~\ref{eq:mce}, the mean
conditional entropy becomes:

\begin{equation}
\label{eq:mce3}
H(Y|\mathbf{x}) = \frac{M - N}{s} H(F(Y)) + \sum_{\mathbf{x} \in
\mathbf{X}:P(\mathbf{x}) > \frac{1}{s}}
P(\mathbf{x}) H(Y|\mathbf{x})\textrm{,}
\end{equation}
\noindent where $F(Y)$ is the probability distribution given by
\[
F(i) \:=\: \left\{\begin{array}{ll} \beta, & \textrm{if } i = 1\\
\frac{1-\beta}{c-1}, & \textrm{if } i = 2, 3 ..., c \end{array} \right.
\textrm{,}
\]

\noindent and $N$ in this case is the number of instances
$\mathbf{x}$ with $P(\mathbf{x}) > \frac{1}{s}$ (more than one observation).

Since $\varepsilon_Y(\mathbf{x}) = 1 - \beta$ when
$P(Y|\mathbf{x}) = \frac{1}{t}$, the $CoD$ with this
penalization is given by:

\begin{equation}
\label{eq:cod3}
CoD_Y(\mathbf{X}) = \frac{\varepsilon_Y - (1 - \frac{(M - N)}{s}\beta -
\sum_{\mathbf{x} \in \mathbf{X}:P(\mathbf{x}) > \frac{1}{s}}
P(\mathbf{x}) max_{y \in Y}P(y|\mathbf{x}))}{\varepsilon_Y}
\end{equation}
\subsection{Classifier design and generalization}
\label{sec:generalization}

After the feature selection using $H$ or $CoD$, the classifier is
designed from the table of conditional probabilities where each row
is a possible instance $\mathbf{x} \in \mathbf{X}$, each column is a
possible class $Y = y$ and each cell of this table represents
$P(Y|\mathbf{X} = \mathbf{x})$. This table is used as a Bayesian
classifier where, for each given instance, the chosen label
$Y = y$ is the one with maximum conditional probability for the considered
instance. In case of instances that have two or more labels of maximum
probability (including non-observed instances), it is possible to
generalize these instances according to some criterion. A commonly used
criterion is the nearest neighbors with some distance metric
\cite{theodoridis99}. We
implemented the nearest neighbors using Euclidean distance. In this
implementation, the nearest neighbors are taken successively. The
occurrences of each label are summed until only one of such
labels has the maximum number of occurrences and may be chosen as the
class to which the considered instance belongs. This featured can be
turned off. In this case, the label is guessed, i.e., chosen randomly
from the labels with maximum number of occurrences (including
non-observed instances).
\section{Software description}
\label{sec:software}

\begin{figure}[ht!]
    \begin{minipage}{0.49\linewidth}
        \subfigure[Upload the biological data]{\includegraphics[width=\linewidth]{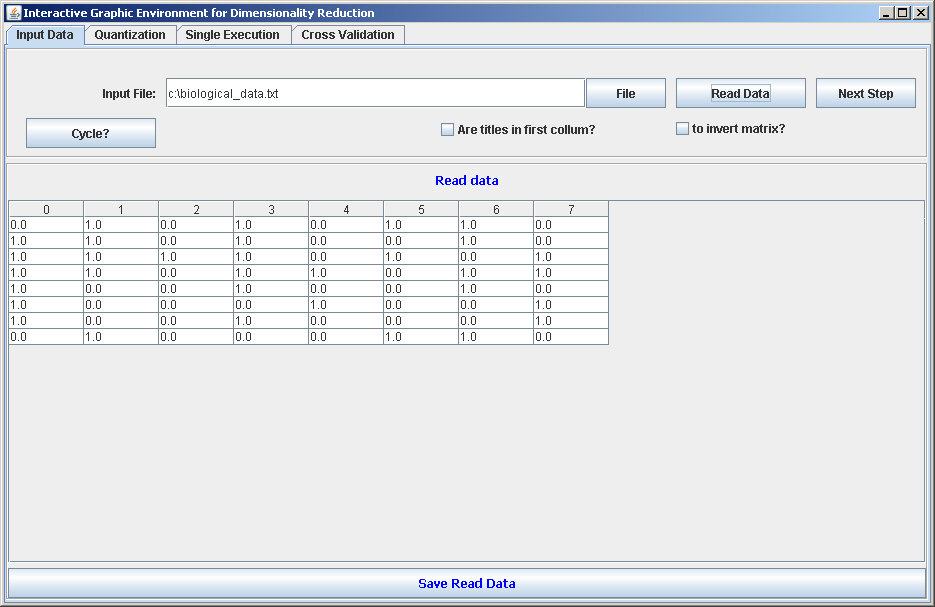}}
    \end{minipage} \hfill
    \begin{minipage}{0.49\linewidth}
        \subfigure[Quantization process]{\includegraphics[width=\linewidth]{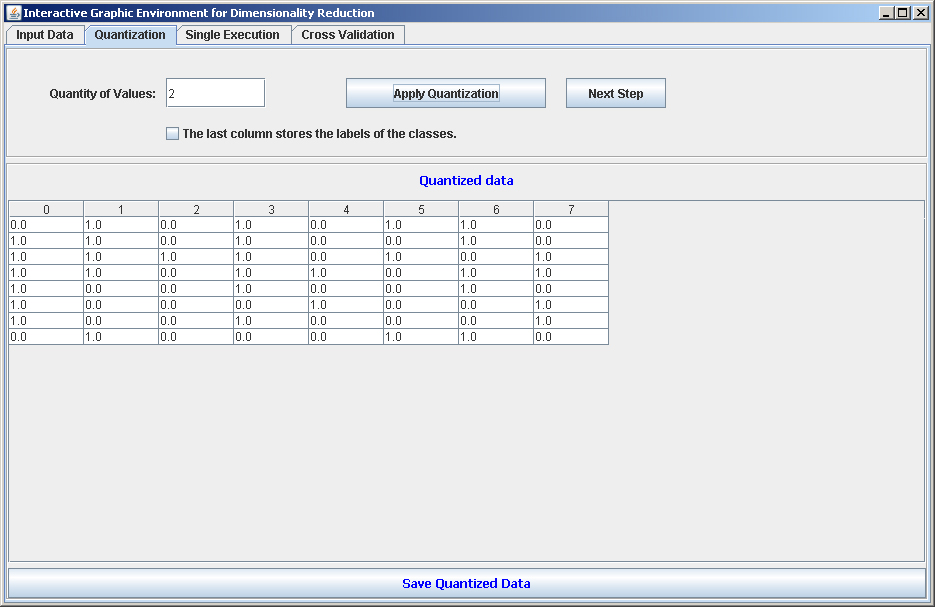}}
    \end{minipage}
    \\
    \begin{minipage}{0.49\linewidth}
        \subfigure[Single execution]{\includegraphics[width=\linewidth]{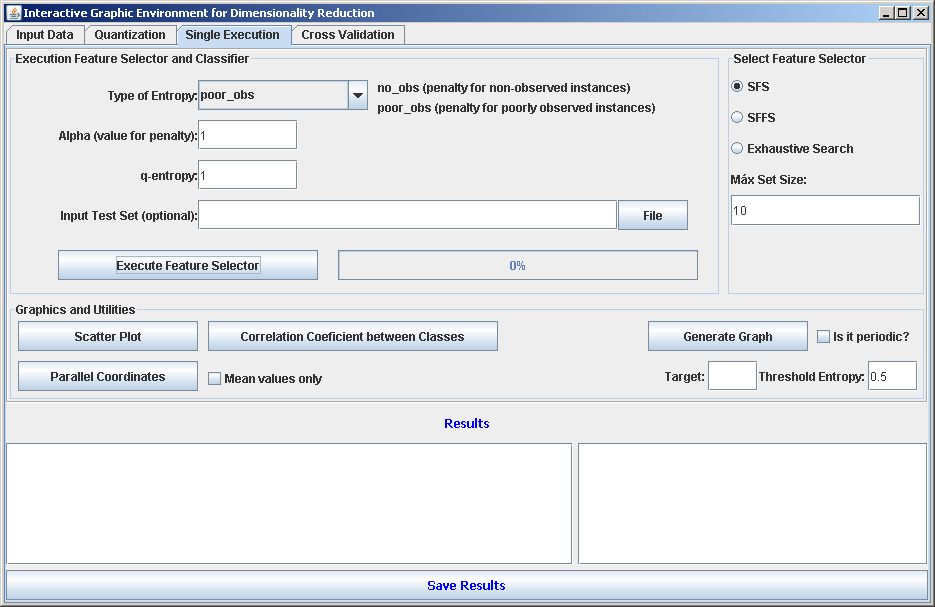}}
    \end{minipage} \hfill
    \begin{minipage}{0.49\linewidth}
        \subfigure[Cross-validation]{\includegraphics[width=\linewidth]{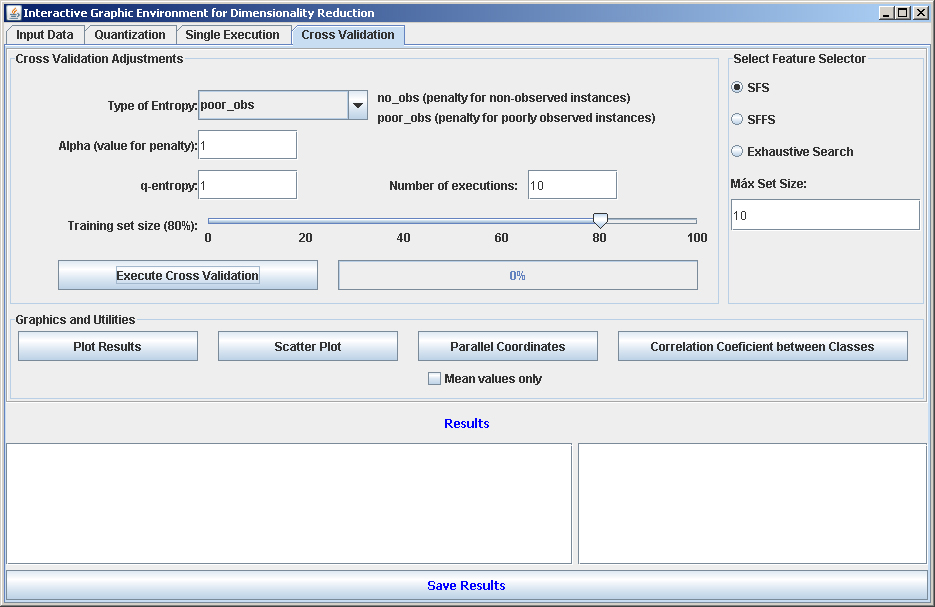}}
    \end{minipage}
    \caption{Application panels.}
    \label{fig:apppanels}
\end{figure}
\begin{figure}[!ht]
\begin{center}
    \begin{minipage}{0.7\linewidth}
        \subfigure[Scatterplot]{\includegraphics[width=\linewidth]{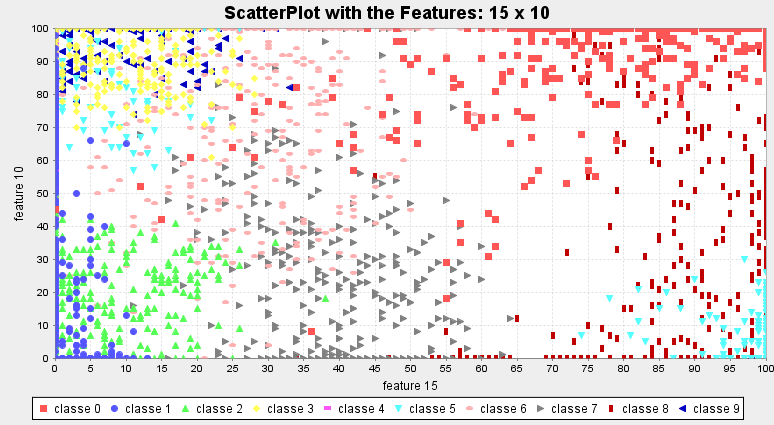}}
    \end{minipage}\\
    \begin{minipage}{0.7\linewidth}
        \subfigure[Parallel coordinates]{\includegraphics[width=\linewidth]{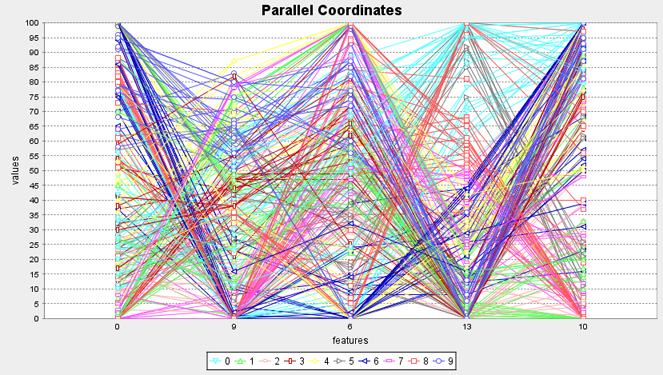}}
    \end{minipage}
    \caption{Examples of scatterplot and parallel coordinates generated by the software.}
    \label{fig:scatter_parallel}
\end{center}
\end{figure}

The software is implemented in Java in order to be executable in
different platforms. It is open source and intended to be continuously
developed in a world-wide collaboration. The software is available at
\url{http://code.google.com/p/dimreduction/}.

There are four main panels: the first panel allows the user to load
the data set (Figure~\ref{fig:apppanels}-a). The second is optional
for the user to define a quantization degree to the data set. The
quantized data may be visualized (Figure~\ref{fig:apppanels}-b).
It is worth noting that some feature selection criteria
like mean conditional entropy or CoD require data quantization to
discrete values. This fact explains the quantization step available in
the software. The data quantization is based on a common rule, searching for
the extreme values (positive and negative) and dividing equally the negative
and positive space considering the number of divisions specified by the
quantization degree parameter.

The next step can be the single execution or cross-validation. The first
one is dedicated to perform single tests (Figure~\ref{fig:apppanels}-c).
It is represented by a panel where the user is able to enter input parameters
such as the feature selection algorithm
(see Section~\ref{sec:feature_selection} for the algorithms
implemented) and the criterion function
(see Section~\ref{sec:criterion_functions} for the criteria
implemented). Other implemented utilities, including the visualization results
of the feature selection, area found in the middle of the panel.
There are three forms to visualize the results:
graphs (Figure~\ref{fig:resultnetrec}),
scatterplot (Figure~\ref{fig:scatter_parallel}-a) and
parallel coordinates (Figure~\ref{fig:scatter_parallel}-b).
The graphs show the connections among different classes, chosen in
feature selection execution, as directed edges between selected vertices. The
parallel coordinates proposed by~\cite{Inselberg85} allows to visualize in adjacent
axes (selected features) similar patterns of behavior in data, visually indicating
how separated are the classes, considering the adjacent features. In the software
application, the features and it and its order to build he parallel coordinates
chart are defined by the user.

The cross-validation panel (Figure~\ref{fig:apppanels}-d) is very similar to the prior.
Cross-validation~\cite{crossvalidation} consists in to divide the
whole data set in two subsets: training and test, mutually
exclusive, and the user can define the size of both sets. The
training set is entered as input to the feature selection
algorithm. The classifier designed from the feature selection and the
joint probability distributions table labels the test set samples. At
the end of the cross-validation process, it is plotted a chart with
the results of each execution, and it is possible to visualize the rate
of hits and its variation along the executions.

Another available option is the generalization of non-observed
instances. With this option selected, the instances of the selected
feature set not present in the training samples are generalized by a
nearest neighbors method~\cite{theodoridis99} with Euclidean distance
(see Section~\ref{sec:generalization} for more details). This method
is also applied to take a decision among classes with tied maximum conditional
probability distributions given a certain instance.

\section{Illustrative Results}
\label{sec:results}

\begin{figure}[!ht]
\begin{center}
    \includegraphics[width=0.7\linewidth]{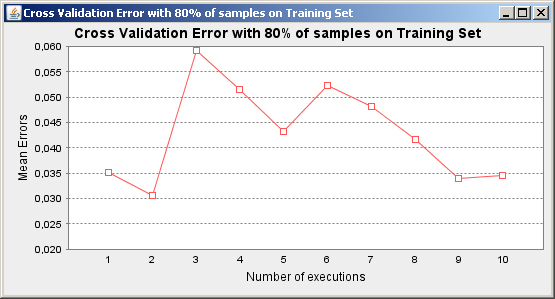}
    \caption{Cross-validation results using 10 executions, 80\% of
      data as training set and 20\% as test set.}
    \label{fig:resultadocv}
\end{center}
\end{figure}

This section presents the results in two main aspects. Initially the software
was applied as feature selection in a biological classification
problem to classify breast cancer cells in two possible classes: benign and
malignant. The biological data used here was obtained
from~\cite{Asuncion+Newman:2007} which has 589 instances and 32 features.
The results shown figure~\ref{fig:resultadocv}, presents very low variations
and high accurate classification achieving 99.96\% of accuracy on average.

\begin{figure}[!ht]
\begin{center}
    \includegraphics[width=0.7\linewidth]{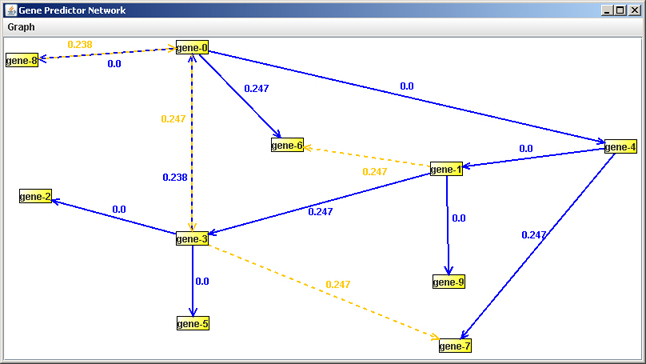}
    \caption{Identified network: dashed lines represent the false
      positives and solid lines the positives. There are no false negatives.}
    \label{fig:resultnetrec}
\end{center}
\end{figure}

The second computational biology problem addressed was gene network recovery.
In this case we used an artificial gene network generated by the approach presented
in~\cite{FRL08}. The parameters used were: 10 nodes, binary
quantization, 20 observations (timestamps), 1 average of edges per
vertex and Random graphs of Erd\"{o}s-R\'{e}nyi as network architecture.
In figure~\ref{fig:resultnetrec}, it is presented the network
recovered. This result did not present false negatives and just few
false positives.

\section{Conclusion}
\label{sec:conclusion}
The proposed feature selection environment allows data analysis using
several algorithms, criterion functions and graphic visualization
tools. Since it is an open-source and multi-platform software, it is
suitable for the user that wants to analyze data and draw some
conclusions about it, as well as for the specialist that has as objective
to compare several combinations of approaches and parameters for
each specific data set or to include more features in the software
such as a new algorithm or a new criterion function. This system
can evolve and include feature extraction methods as well, not
limited only to feature selection methods.

The environment can be used in many pattern recognition
applications, although the main concern is with Bioinformatics
tasks, especially those involving high-dimensional data
(large number of genes, for example) with small number
of samples. Even users not familiar with programming
are allowed to manipulate the software in an easy way,
just by clicking to select file inputs, quantization,
algorithms, criterion functions, error estimation methods
and visualization of the results. The environment is
implemented as ``wizard style'', i.e., it has tabs
delimiting each procedure.

This software opens a great space for future works. The next step
consists in the implementation of other classical feature selection
algorithms (e.g. GSFS and PTA \cite{theodoridis99,Somol}),
criterion functions (e.g. based on distances between classes
\cite{theodoridis99}), error estimation methods (e.g. Leave-one-out
and Bootstrap) and then the inclusion of classical methods of feature
extraction (e.g. PCA~\cite{Jolliffe}).

\section*{Acknowledgement}
This work was supported by FAPESP, CNPq and CAPES.

\bibliographystyle{unsrt}
\bibliography{dimreduction-full}
\end{document}